\documentclass{article}

\usepackage[nonatbib,preprint]{neurips_data_2024}
\usepackage[numbers]{natbib} 
\usepackage{graphicx}
\usepackage{enumitem}
\usepackage{pifont}
\usepackage{makecell}
\usepackage{multirow}
\usepackage[table]{xcolor}%
\usepackage{tcolorbox}
\usepackage{booktabs}

\usepackage[utf8]{inputenc} %
\usepackage[T1]{fontenc}    %
\usepackage[pagebackref,breaklinks,colorlinks,allcolors=blue]{hyperref}
\usepackage{url}            %
\usepackage{booktabs}       %
\usepackage{amsfonts}       %
\usepackage{nicefrac}       %
\usepackage{microtype}      %
\usepackage{xcolor}         %

\title{VSP: Assessing the dual challenges of perception and reasoning in spatial planning tasks for VLMs}

\author{%
  Qiucheng Wu${^1}$, Handong Zhao${^2}$, Michael Saxon${^1}$,\\ \textbf{Trung Bui}${^2}$, \textbf{William Yang Wang}${^1}$, \textbf{Yang Zhang}${^3}$, \textbf{Shiyu Chang}${^1}$ \\
  $^1$UC Santa Barbara, $^2$Adobe Research, $^3$MIT-IBM Watson AI Lab\\
  \texttt{qiucheng@ucsb.edu} \\
}

\begin{document}

\maketitle

\begin{abstract}
Vision language models (VLMs) are an exciting emerging class of language models (LMs) that have merged classic LM capabilities with those of image processing systems. 
However, the ways that these capabilities combine are not always intuitive and warrant direct investigation. 
One understudied capability in VLMs is \textit{visual spatial planning}---the ability to comprehend the spatial arrangements of objects and devise action plans to achieve desired outcomes in visual scenes. 
In our study, we introduce \textbf{VSP}, a benchmark that
1) evaluates the spatial planning capability in these models in general, and 2) breaks down the visual planning task into finer-grained sub-tasks, including perception and reasoning, and measure the LMs capabilities in these sub-tasks. 
Our evaluation shows that both open-source and private VLMs fail to generate effective plans for even simple spatial planning tasks. 
Evaluations on the fine-grained analytical tasks further reveal fundamental deficiencies in the models’ visual perception and bottlenecks in reasoning abilities, explaining their worse performance in the general spatial planning tasks.
Our work illuminates future directions for improving VLMs' abilities in spatial planning.
Our benchmark is publicly available at \url{https://github.com/UCSB-NLP-Chang/Visual-Spatial-Planning}.
\end{abstract}

\section{Introduction}
The rapid advancement of large language models has driven considerable growth in their capabilities to produce fluent text in many domains, generating outputs exhibiting potential ``reasoning'' and ``understanding'' abilities \cite{touvron2023llama,bi2024deepseek,jiang2024mixtral,brown2020language}. %
Recently, vision language models (VLMs) have advanced on LMs through additional training on native image inputs, to achieve impressive performance generating text describing and relating to input images
\cite{achiam2023gpt,liu2024visual,team2023gemini,awadalla2023openflamingo,alayrac2022flamingo}, 
 with applications in image captioning, visual question answering, visual reasoning, and others \cite{ying2024mmt,yang2024exploring,shao2023prompting,zheng2023ddcot}. 
The swift evolution of VLMs has enabled them to tackle increasingly sophisticated tasks that require multiple emerging abilities in complex scenarios. 
However, as model capabilities and deployment needs advance, the challenges in usefully evaluating them grow in kind.

Planning is a fundamental capability in intelligent systems that is particularly contested in LMs \cite{valmeekam2023planning}, and is {understudied in VLMs}.
\textit{Visual spatial planning} refers to the task of comprehending the spatial arrangement of objects in a scene and designing action plans to achieve a desired outcome. 
For example, the classical maze problem can be considered a visual planning task, where an agent is given an input image describing the maze environment and is asked to produce a viable path to navigate the player from the starting position to the goal. 
This task requires two capabilities: \emph{image perception}, which enables the agent to understand the objects, environment and spatial relations present in the image, and \emph{reasoning}, which enables the agent to perform strategic decision-making.

Visual spatial planning is an important capability in many potential applications for VLMs, such as navigating in complex environments with autonomous driving~\cite{tian2024drivevlm,ma2023dolphins} or manipulating objects with robotic hands~\cite{chang2023lgmcts,hu2023look}. Though there have been increasingly more benchmarks to evaluate the vision processing capabilities of VLMs, few current benchmarks systematically evaluate their capability to perform visual spatial planning tasks. As shown in Table~\ref{tab:comparison-baselines}, existing benchmarks mostly focus on VLMs' ability to understand image content and perform visual logic reasoning~\cite{fu2023mme,yue2023mmmu,wang2024measuring}; however, they often overlook the ability to comprehend the spatial arrangements of entities within images and to devise spatial action plans based on practical restrictions in the visual environment. As a result, two research questions are left unanswered:  \ding{182} How performant are VLMs in performing visual planning tasks?  \ding{183} What are the bottleneck capabilities, \emph{e.g.}, perception or reasoning, that limit the performance of VLMs in the visual planning tasks?

To this end, we introduce \underline{V}isual \underline{S}patial \underline{P}lanning (VSP), a benchmark specifically designed to evaluate the spatial planning capabilities of VLMs. As illustrated in Figure~\ref{fig:intro-maze} and Figure~\ref{fig:intro-blocks}, the VSP benchmark is developed from classical maze navigation and block-moving games, where the entire environment is fully observable in the input images. In this benchmark, the VLMs are required to interpret the visual inputs, deduce the consequences of each action, and execute the designated tasks accordingly. To comprehensively evaluate the fine-grained capabilities needed for the visual spatial planning, our VSP includes 4.4K questions in 10 meticulously designed tasks that feature both simulated and photo-realistic visual environments. In addition to testing end-to-end spatial planning performance, these tasks further evaluate essential individual capabilities needed for performing visual planning, such as image perception and reasoning.

We apply the VSP benchmark to evaluate existing state-of-the-art VLMs, including both open-source and private models. Surprisingly, we find that even the most competitive VLMs sometimes struggle in performing the simplest visual planning tasks, such as a 3x3 maze problem or an one-step block-moving task. 
Our fine-grained capability analysis further reveals that existing VLMs have flaws in reasoning and bigger bottlenecks in perception.
We believe the VSP benchmark highlights critical weaknesses in current VLMs and sheds light on future directions for enhancing their spatial understanding and planning capabilities.

\begin{table*}[t]
\centering
\caption{Comparison with representative existing benchmarks.}
\vspace*{0.1in}
\resizebox{1.0\linewidth}{!}{
\begin{tabular}{lll}
\toprule
 \textbf{Name} & \textbf{Tasks Description}  & \textbf{Keywords}\\ 
\midrule 
{\footnotesize \textbf{MME}~\cite{fu2023mme}} & {\footnotesize Image content understanding, reasoning} &{\footnotesize perception, reasoning}  \\
{\footnotesize \textbf{MMMU}~\cite{yue2023mmmu}} & {\footnotesize College-level knowledge reasoning} &{\footnotesize multi-discipline knowledge, reasoning}  \\
{\footnotesize \textbf{MathVision}~\cite{wang2024measuring}} & {\footnotesize Math problems with visual contexts} &{\footnotesize mathematical reasoning}  \\
{\footnotesize \textbf{SeedBench}~\cite{li2023seed}} & {\footnotesize Comprehension of scene \& instance in image} &{\footnotesize perception, reasoning, spatial relation}  \\
{\footnotesize \textbf{MM-Vet}~\cite{yu2023mm}} & {\footnotesize General problems that need integrated abilities} &{\footnotesize perception, reasoning, spatial relation}  \\
\midrule
{\footnotesize \textbf{VSP}} & {\footnotesize \makecell[l]{Understand \& extract\\ \textit{spatial} info and \textit{plan accordingly}}} &{\footnotesize \makecell[l]{\textit{Spatial} planning,\\ \textit{Spatial} perception, reasoning}}  \\
\bottomrule

\end{tabular}
}
\label{tab:comparison-baselines}
\end{table*}

\section{Related Work}
\subsection{General planning in LMs}
Planning has been a central focus of research in AI. Traditional work in AI planning includes using formal languages to represent and solve planning problems~\cite{aeronautiques1998pddl}, and developing algorithms like dynamic programming and reinforcement learning to explore environments and formulate viable plans~\cite{guo2014deep,sutton1991planning}. 
While these works mostly focus on predefined and restricted environments, recently, with the advancement of LMs, it has become intriguing to study whether LMs, with the potential to be general intelligent agents, can perform planning in different settings and environments~\cite{kambhampati2024can,kambhampati2024llms,stechly2024self}.
Many works explore the best ways to activate the planning capabilities of LMs, including divide and conquer~\cite{wei2022chain,yao2024tree,shen2024hugginggpt,yao2022react}, grounding outputs in admissible actions~\cite{ahn2022can,hazra2024saycanpay}, retrospecting and refining~\cite{shinn2024reflexion,madaan2024self}, and leveraging external tools~\cite{guan2023leveraging,ruan2023tptu}.
Meanwhile, with the increasing capabilities of LMs, growing research efforts are now dedicated to benchmark their planning capabilities in various complex environments~\cite{wu2023smartplay,xie2024travelplanner,valmeekam2022large}.

\subsection{Spatial and visual planning in LMs}
Many general planning tasks in LMs involve understanding visual environments and comprehending spatial information. In robotics and embodied agent studies, LMs play a crucial role in grounding visual entities with references in open-domain instructions and formulating plans based on spatial constraints. Consequently, they are increasingly used in physically grounded scenarios such as object rearrangement~\cite{chang2023lgmcts,hu2023look}, cooking~\cite{joublin2023copal,sakib2023cooking}, and navigation~\cite{hazra2024saycanpay,ahn2022can}. 
LMs are also used in AIGC to propose spatial arrangements of entities following instructions~\cite{feng2024layoutgpt}.
While realistic planning tasks align with real needs, their complexity and expansive action spaces limit the analysis of LMs' detailed planning capabilities.
Therefore, research also focuses on LMs' planning in simulated environments and games.
For example, \textit{mystery blocksworld} is a dynamically generated set of blocksworld tasks to test generalization in LMs~\cite{valmeekam2023planning}. Additionally, many text games have been introduced to test LMs' abilities in spatial understanding and imagination~\cite{shridhar2020alfworld,wu2023smartplay,yang2023dawn,aghzal2023can}. However, most of these studies transform visual information into text inputs, thus not directly measuring LMs' visual abilities.

\subsection{Benchmarks for VLMs}
VLMs have inherited and advanced many intriguing features from text-only LMs~\cite{yang2023dawn,qi2023gemini}.
Benchmarks for VLMs have rapidly emerged to evaluate performance in areas such as image content understanding~\cite{fu2023mme,cha2024visually}, perception~\cite{ge2023mllm,tong2024eyes}, knowledge~\cite{yue2023mmmu,wang2024measuring,lu2023mathvista}, and reasoning~\cite{fu2023mme,yue2023mmmu,liu2023hallusionbench}.
Recently, there are also emerging benchmarks focusing on the capability of understanding multiple images in long context and complex realistic environments~\cite{song2024milebench,haystack}.
While these benchmarks successfully quantify VLMs' abilities in many fields, their capabilities in spatial understanding and reaction are relatively under-explored.
Some benchmarks cover spatial relations understanding~\cite{li2023seed,yu2023mm}, but often overlook the ability to devise complex spatial action plans based on visual environment constraints.
We focus on \textit{visual spatial planning} - the ability to comprehend spatial arrangements of objects and devise action plans to achieve specific outcomes.
We fill the gap in benchmarking VLM abilities for visual spatial planning and highlight future directions for improving VLMs towards models with general intelligence.

\begin{figure*}[t!]
\begin{center}
\includegraphics[width=\textwidth]{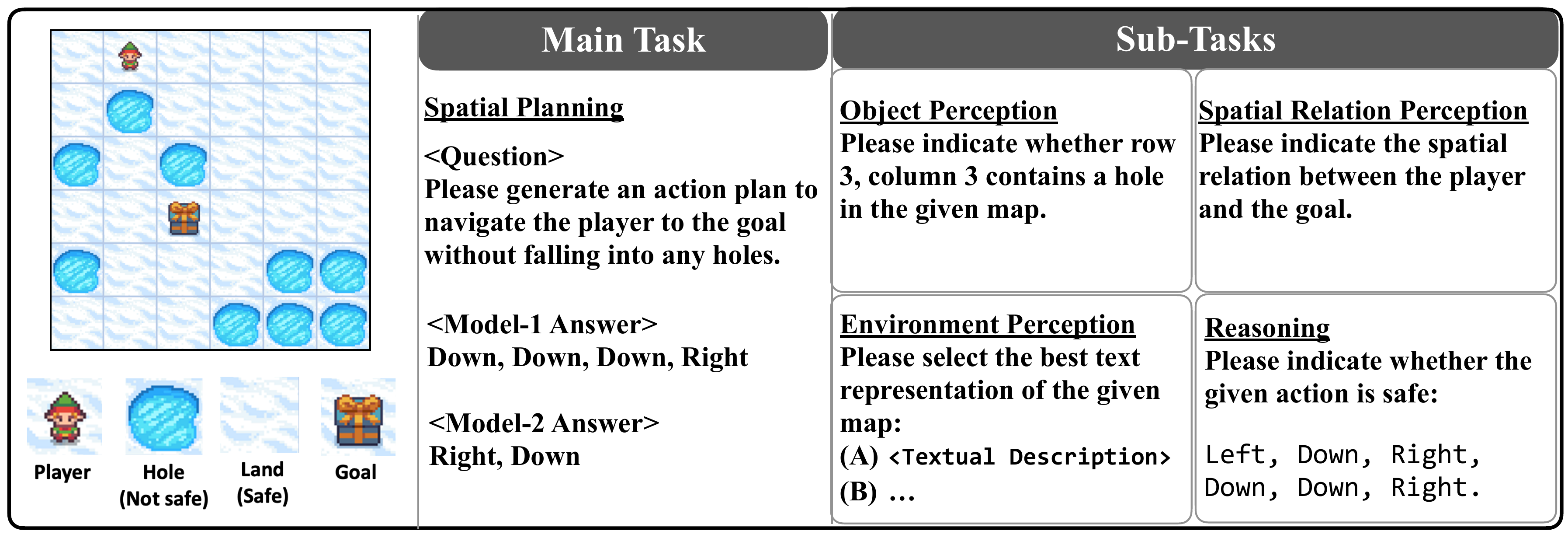}
\end{center}
\vskip -0.1in
\caption{Overview of the Maze Navigation scenario.}
\label{fig:intro-maze}
\end{figure*}

\section{The Visual Spatial Planning Benchmark}
\subsection{Overview of the Benchmark}\label{overview-bench-subsec}
In this benchmark, our objectives are two-fold: \ding{182} quantify the visual spatial planning capabilities of current VLMs; and \ding{183} uncover current capability bottlenecks that limit the effectiveness of VLMs in visual spatial planning tasks.
While the first objective can be achieved through direct measurements on corresponding tasks, the second objective requires more careful benchmark design. Specifically, performing spatial planning in visual environments requires a series of cohesive steps. For example, to generate an accurate path to navigate a player to a goal, an agent needs to be able to correctly view and understand the visual map, reason to find which actions are safe or dangerous, and come up with a detailed plan to achieve the goal. Each of these steps could be challenging for a developing VLM, and understanding which of these subtasks challenge them most will drive future improvement.

To this end, we propose the Visual Spatial Planning (VSP) benchmark, with the objective of measuring and diagnosing the capabilities of VLMs in producing accurate spatial plans in visual environments. The VSP benchmark consists of two scenarios: \ding{182} the simulated \textbf{Maze Navigation scenario}, whose main task is to move a game character through a maze, and \ding{183} the photo-realistic \textbf{Blocks World scenario}, whose main task is to move blocks from a starting configuration to a goal configuration.
In each scenario, in addition to the main task, VSP introduces four sub-tasks that focus on the individual capabilities needed for the main task:

$\bullet$ \textbf{T1. Single Object Perception} -- Determine the characteristics of a single object; 

\vspace{-0.02in}
$\bullet$ \textbf{T2. Spatial Relation Perception} -- Determine the relative positions of two objects; 

\vspace{-0.02in}
$\bullet$ \textbf{T3. Environment Perception} --  Find textual descriptions that describe the visual environment;

\vspace{-0.02in}
$\bullet$ \textbf{T4. Spatial Reasoning} -- Determine the consequence of a series of actions or moves. 

The sub-task details are designed specific to each scenario. Furthermore, to demonstrate the model’s performance under different levels of environmental complexity, we establish progressive difficulty settings for each task, which are measured by parameters such as map size, minimum required number of actions, \emph{etc.} 
We provide the details of task statistics, \emph{i.e.,} total number of problems, in appendix~\ref{detail-task}. In what follows, we will introduce each scenario in detail, as well as the data curation and the task creation processes.

\begin{figure*}[t!]
\begin{center}
\includegraphics[width=\textwidth]{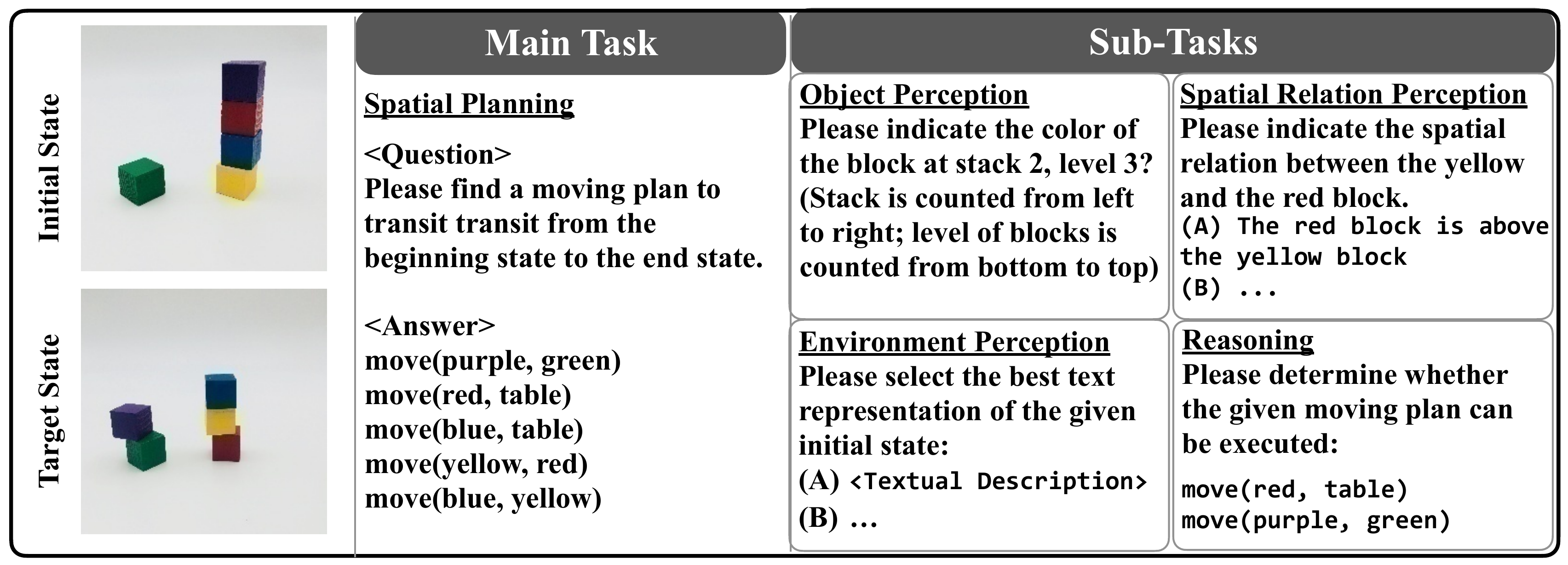}
\end{center}
\vskip -0.1 in
\caption{Overview of the Blocks World scenario.}
\label{fig:intro-blocks}
\end{figure*}
\subsection{The Maze Navigation Scenario}
The maze navigation scenario is inspired by the popular implementation~\cite{brockman2016openai} of a fully observable path-finding problem. As depicted in Figure~\ref{fig:intro-maze} left, it simulates a classical grid world environment with a designated start and goal position, where part of the grids contain obstacles (the ``holes'') and cannot be passed through. 

The main spatial planning task and the four sub-tasks are defined as follows:

$\bullet$ \textbf{Main Task} (Spatial Planning) -- Generate a safe path to navigate from the start grid to the goal;

\vspace{-0.02in}
$\bullet$ \textbf{T1} (Single Object Perception) -- Determine if a specified grid is safe;

\vspace{-0.02in}
$\bullet$ \textbf{T2} (Spatial Relation Perception) -- Find spatial relations between the player and the goal;

\vspace{-0.02in}
$\bullet$ \textbf{T3} (Environment Perception) -- Find the textual description that fits the visual environment;

\vspace{-0.02in}
$\bullet$ \textbf{T4} (Spatial Reasoning) -- Determine the consequence of a given action series.

An example of input image and questions is demonstrated in Figure~\ref{fig:intro-maze}. Each task is equipped with progressive adjusted difficulty settings to evaluate the model's capability under various circumstances. For \textbf{Main Task} and \textbf{T1}-\textbf{T3}, the difficulties are measured by the size of the map, ranging from 3x3 to 8x8, where a larger map introduces more challenges in correctly perceiving objects and planning accordingly. For task \textbf{T4}, since a longer path naturally introduces more challenges for reasoning, we adopt path length ranging from 1 to 9 as the difficulty measure. Please refer to Appendix~\ref{detail-task} for the complete example of the question and answer in each task.

\subsection{The Blocksword Scenario}
\textit{Blocksworld} is a widely-adopted planning problem~\cite{valmeekam2022large,hao2023reasoning,gokhale2019blocksworld}. As depicted in Figure~\ref{fig:intro-blocks} left, in this scenario, the agent is given images containing sets of blocks in unique colors. These blocks are stacked vertically, forming multiple stacks on the table. The agent is asked to turn the blocks from initial state to target state through a series of moving actions. For each action, the agent can only move the top block of any stack, providing it is moved to either the table or the top of another stack.

Similarly, the main spatial planning task and the four sub-tasks are defined as follows:

$\bullet$ \textbf{Main Task} (Spatial Planning) -- Form a moving plan to achieve the target state of block arrangement;

\vspace{-0.02in}
$\bullet$ \textbf{T1} (Single Object Perception) -- Determine the color of the block at a specific position;

\vspace{-0.02in}
$\bullet$ \textbf{T2} (Spatial Relation Perception) -- Determine the spatial relation between two blocks;

\vspace{-0.02in}
$\bullet$ \textbf{T3} (Environment Perception) -- Find the text representation that fits the visual environment;

\vspace{-0.02in}
$\bullet$ \textbf{T4} (Spatial Reasoning) -- Determine the consequence of a given moving plan.

An example of input image and questions is demonstrated in Figure~\ref{fig:intro-blocks}. Similar to the maze navigation scenario, each task is equipped with progressive adjusted difficulty. Specifically, in \textbf{Main Task} and \textbf{T4}, the difficulties are measured by the number of actions involved, ranging from 1 to 7, which quantifies the complexity of the action plan. On the other hand, for tasks \textbf{T1}-\textbf{T3}, which focus on perception, the difficulty is measured by the number of blocks presented in the image, ranging from 3 to 5. Please refer to Appendix~\ref{detail-task} for the complete example of the questions in each task.

\begin{figure*}[t!]
\begin{center}
\includegraphics[width=\textwidth]{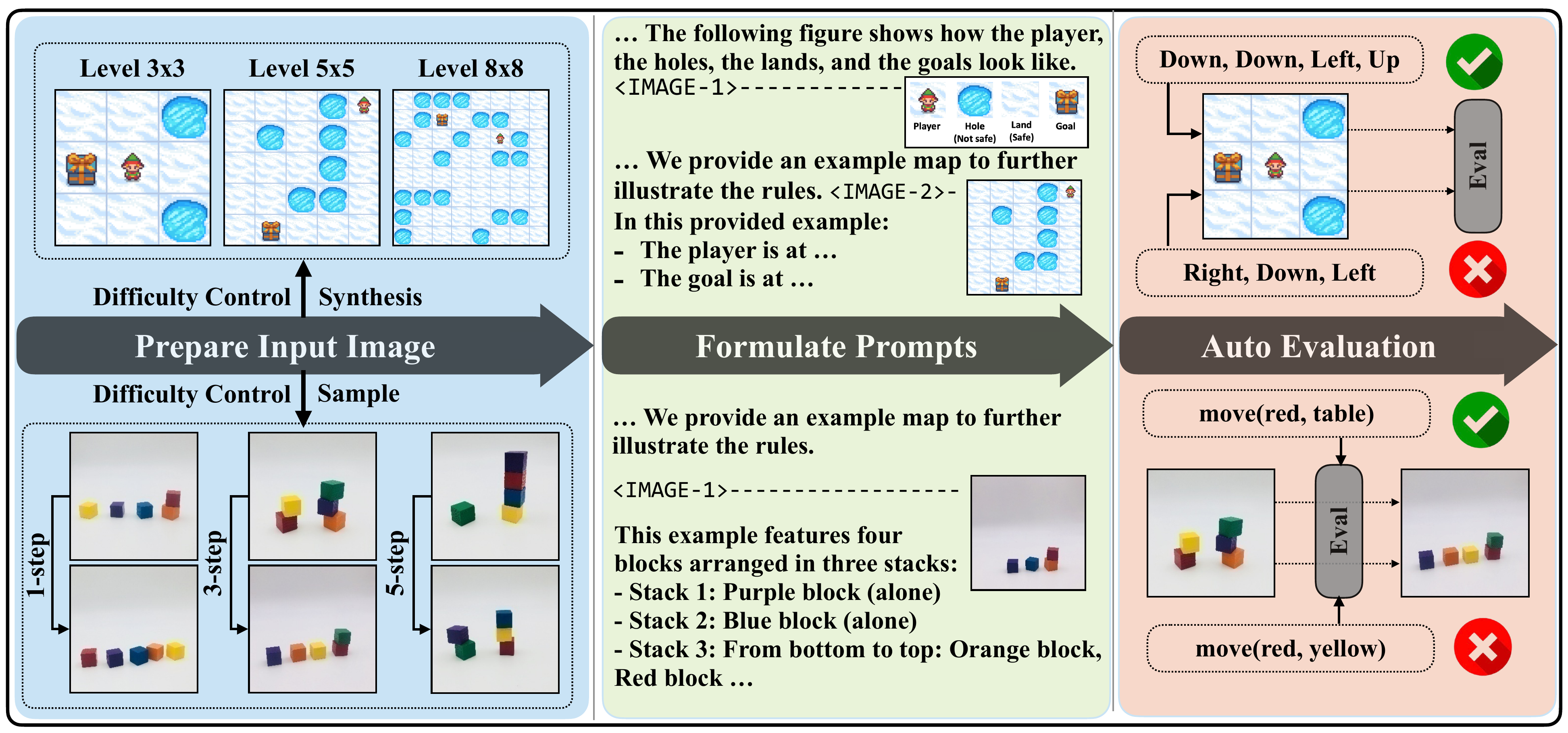}
\end{center}
\vskip -0.08 in
\caption{Benchmark creation process. \textbf{Left:} We prepare input images that fulfill the task requirements with different difficulties. \textbf{Mid:} We formulate input prompts for each task. The input prompts consists of interleaved texts and images. \textbf{Right:} We develop automatic evaluation process for each task.}
\label{fig:bench-form-process}
\end{figure*}
\subsection{Benchmark Creation Process}\label{main-text-benchmark-creation}
Figure~\ref{fig:bench-form-process} demonstrates our 3-stage general process for benchmark creation. 

First, in the left panel of Figure~\ref{fig:bench-form-process}, we prepare the input images used for each task and scenario. In the maze navigation scenario, we generate input maps using the OpenAI Gym package~\cite{brockman2016openai}, with modifications to ensure that the positions of the player, the goal, and the holes are all randomly generated. In the blocksworld scenario, we sample pairs of images from the BIRD dataset~\cite{gokhale2019blocksworld}, ensuring there is at least one viable plan to move the blocks from the initial state to the target state. 
The images are prepared conditional on different levels of difficulty.

Second, in the center panel of Figure~\ref{fig:bench-form-process}, we formulate input prompts for each task. The prompt consists of interleaved text and images to provide sufficient information. For example, for maze navigation, we include images to show the appearance of elements in the map and provide example maps to better illustrate how the models should interpret the map. We invite native speakers to refine the prompts so that they accurately describe the task requirements. These prompts are demonstrated in Appendix~\ref{detail-task}.

Finally, in the right panel of Figure~\ref{fig:bench-form-process}, we evaluate the performance of VLMs under each task. It is worth noting that the answer for each task is often not unique. For example, in the blocksworld scenario, there can be many ways to move the blocks to reach the target state. As such, we develop scripts to automatically evaluate the answers for each task.

In addition to the steps above, some tasks require extra steps to construct meaningful questions, candidates, and answers, such as prompt design or example filtering. For example, in task 5 of the blocksworld scenario, the input actions must cover various valid/invalid movements, requiring filtering and balancing. The detailed steps we followed to create each task set are provided in Appendix~\ref{detail-task}. We release all images, texts, and scripts to facilitate replication and scaling.
\begin{table}[!t]
\centering
{ 
    \caption{{Zero-shot success rates for the spatial planning task, at various difficulty levels. Maze navigation difficulty levels represent the maze's square grid length. Blocksworld difficulty levels correspond to the minimum number of steps to a solution. Results better than 30\% are \textbf{bolded}.}}
    \footnotesize
    \resizebox{\linewidth}{!}{
\begin{tabular}{@{}lccccccccccc@{}}
\toprule
     & \multicolumn{6}{c}{\textsc{Maze Navigation}}
       & \multicolumn{4}{c}{\textsc{Blocksword}} & {\textsc{Overall}}   \\
     \cmidrule(lr){2-7}
     \cmidrule(lr){8-11}
     \small \textbf{Difficulty level} & {\small{\textbf{3}}} & {\small{\textbf{4}}} & {\small{\textbf{5}}} & {\small{\textbf{6}}} & {\small{\textbf{7}}} & {\small{\textbf{8}}} & {\small{\textbf{1}}} & {\small{\textbf{3}}} & {\small{\textbf{5}}} & {\small{\textbf{7}}} & \\
     \midrule
{\small Gemini\cite{team2023gemini}} &  \textbf{0.31} &	0.26	&0.15&	0.06&	0.14&	0.10 & 0.10 &	0.14 &	0.00	& 0.01 & 0.13   \\
{\small GPT-Vision\cite{achiam2023gpt}  } &   \textbf{0.55} &	\textbf{0.36} &	0.27&	0.13&	0.17&	0.10 & \textbf{0.50} &	0.17	& 0.03 & 	0.00  & 0.23  \\ 
{\small Claude-3\cite{anthropic2024claude}   } &   \textbf{0.52} &	\textbf{0.33} &	0.16&	0.15&	0.16&	0.09&   0.12 & 0.03 & 0.00 & 0.00 & 0.16        \\ 
{\small GPT-4o\cite{GPT-4o}  } & \textbf{0.68}	& \textbf{0.58} & 	\textbf{0.35} &	0.24	 & 0.18 & 	0.23 &    \textbf{0.71} &	\textbf{0.33} &	0.12 &	0.03 &\textbf{0.35}        \\ 
\midrule
{\small LLaVA\cite{liu2024visual}     } &  0.03 &	0.03 & 0.02	& 0.08 &	0.09 &	0.04& 0.04 & 	0.01 & 	0.00 & 	0.00 & 0.04     \\
{\small InternLM\cite{dong2024internlm}  } &  0.27 &	0.16 & 	0.06 &	0.05 &	0.04	& 0.07 &  0.10 & 	0.03 & 	0.00 &	0.00 & 0.08   \\
{\small InternLM-VL\cite{dong2024internlm}                   } &  0.15	& 0.14 &	0.08 &	0.04 &	0.02 &	0.05 & 0.02 & 0.00 & 0.00 & 0.00 & 0.05   \\ 
{\small InstructBLIP\cite{dai2024instructblip}                  } & 0.00 & 0.00 & 0.00 & 0.00 & 0.00 & 0.00 & 0.00 & 0.00 & 0.00 & 0.00 & 0.00     \\
{\small SPHINX\cite{lin2023sphinx}                   } &   0.11 &	0.08 &	0.05 &	0.02 &	0.04 &	0.03 & 0.07 & 0.06 & 0.01 & 0.00 & 0.05     \\ \bottomrule
\end{tabular}
}
    \label{tab:spatial-planning-zero-shot}
}
\end{table}
\section{Experiments}
In this section, we present evaluation results of state-of-the-art VLMs under our main tasks and sub-tasks. Our goal is to answer the following research questions: \ding{182} How well can state-of-the-art VLMs perform in the visual spatial planning tasks?
\ding{183} What are the bottleneck capabilities that limit the VLMs in visual spatial planning tasks? 

\subsection{Baselines}
We evaluate various representative VLMs including both private and open-source models. 

We cover the following \textit{private models}: 
\ding{172} Gemini~\cite{team2023gemini} has demonstrated remarkable capabilities in image understanding and reasoning. We adopt \texttt{Gemini-1.0-Pro-Vision} in our experiments \footnote{The latest \texttt{Gemini-1.5-Pro-Vision} currently has a daily request limits of 50. Therefore, we did not include its evaluation.}. 
\ding{173} GPT-4 Turbo with vision~\cite{achiam2023gpt} inherent strong text understanding capabilities from GPT-4 and is equipped with vision capabilities. We use \texttt{turbo-2024-04-09} for evaluation. 
\ding{174} Claude-3~\cite{anthropic2024claude} is a family of VLMs strong at advanced reasoning and vision analysis. We adopt \texttt{claude-3-sonnet-20240229}, the default model used in chat interface and has comparable speed \& cost with GPT Vision. 
\ding{175} GPT-4o~\cite{GPT-4o} is a recently released multimodal LM with one of the most advanced abilities in processing combination of text, audio, and image outputs. We adopt \texttt{gpt-4o-2024-05-13} in experiments.

We cover the following \textit{open-source models}: 
\ding{176} LLaVA~\cite{liu2024visual} performs instruction tuning on LLaMA and projects input image into text embedding space through pre-trained CLIP visual encoder~\cite{radford2021learning}. We adopt \textsc{llava-v1.6-vicuna-7b} for evaluation. 
\ding{177} InternLM-XComposer2~\cite{dong2024internlm} enhances ability to understand free-form text-image composition and surpasses GPT-4V in several tasks. The latest released checkpoints include \texttt{internlm-xcomposer2-7b} and \texttt{internlm-xcomposer2-vl-7b}, with the former focusing on general text-image composition and the latter focusing on VL benchmarks. We adopt both for evaluation.
\ding{178} InstructBLIP~\cite{dai2024instructblip} is a popular VLM based on pre-trained BLIP-2~\cite{li2023blip} model. We adopt \texttt{blip2-t5-instruct-flant5xxl} for evaluation. 
\ding{179} SPHINX~\cite{lin2023sphinx} unfreezes the LLM during pre-training to enhance cross-model alignment. We adopt \texttt{SPHINX-v2-1k} for evaluation.
Additionally, we attempt to perform measurements on the latest CogVLM2~\cite{wang2023cogvlm} fine-tuned on LLaMA-3~\cite{llama3}. However, since its current codebase only supports single-image input, we do not include its results.
For all open-source models, we use their public released checkpoints, codes, and hyperparameter choices.

\subsection{Main Task (Spatial Planning) Evaluation}
First, we present the main task evaluation results for both the maze and the Blocksword scenarios, which reflect the general spatial planning capabilities of existing VLMs.
All the evaluation in this section is conducted under zero-shot setting without any fine-tuning or in-context learning. Evaluations with in-context learning and fine-tuning are presented in Sections~\ref{subsec:icl} and \ref{subsec:finetuning}.

The performance is demonstrated in Table~\ref{tab:spatial-planning-zero-shot}. Each column represents a difficulty level, which is measured by the size of the map (3 represents 3x3 maps) in Maze Navigation and by the minimum number of steps in Blocksword. From the table, we summarize our findings as follows:

\textbf{VLMs have considerable room for improvement in spatial planning tasks.} We observe that both private and open-source models exhibit sub-optimal performance in various scenarios. In particular, open-source models face significant challenges and rarely succeed in these tasks. Besides, even the most capable private models could frequently make mistakes on relatively simple tasks, such as those involving a 3x3 size map or a single-step block moving task. Considering that these tasks would be simple for humans, the VSP benchmark poses a substantial challenge to VLMs, illustrating that current VLMs have considerable potential for improvement in spatial planning tasks.

\textbf{Quick performance decay as difficulty increases.} We observe a significant drop in the success rates of VLMs as task difficulty escalates. For example, GPT-Vision may achieve a success rate of over 50\% on 3x3 size maps, but this plummets to just 10\% on 8x8 maps. Analyzing the impact of increased difficulty, we identify two major challenges for the models: 
\underline{First}, increasing size of the map in maze navigation scenario could make it difficult for the model to accurately \textit{perceive} the positions of elements within the map. 
\underline{Second}, the increase in both map size and the number of steps required for moving blocks heightens the challenge for the model to \textit{reason} deeply through the entire path and devise a complete, viable solution.
In the following experiments, we focus on these two factors and provide in-depth analysis with subsequent tasks.

\textbf{Challenges in open-source models.} Finally, we note that open-source models often face challenges when evaluating on these tasks. We identify two main factors. \ding{172} \underline{Context length:} Open-source models typically have significantly shorter context windows compared to private models. Besides, image embeddings can occupy many tokens. Thus, these models may not have enough capacity to understand the complete inputs.
For example, \textsc{llava-v1.6-vicuna-7b} is trained with a maximum context window of 2048 tokens, while each image consumes 576 tokens. Consequently, when fed with multiple images and relatively long texts in our tasks, the total token length may surpass training, resulting in poor performance. 
\ding{173} \underline{Multiple image input}: Our tasks require the model to understand multiple images interleaved with text inputs, whereas many open-source models are only trained with single-image inputs, with the image positioned at the start of the input. 
To further explore their potential in our tasks, we assess their performance after training on our inputs in Section~\ref{subsec:finetuning}. Meanwhile, we suggest that future open-source models could consider increasing their context length and reducing restrictions on input formats to address complex and realistic tasks effectively.

\begin{figure*}[t]
\begin{center}
\includegraphics[width=\textwidth]{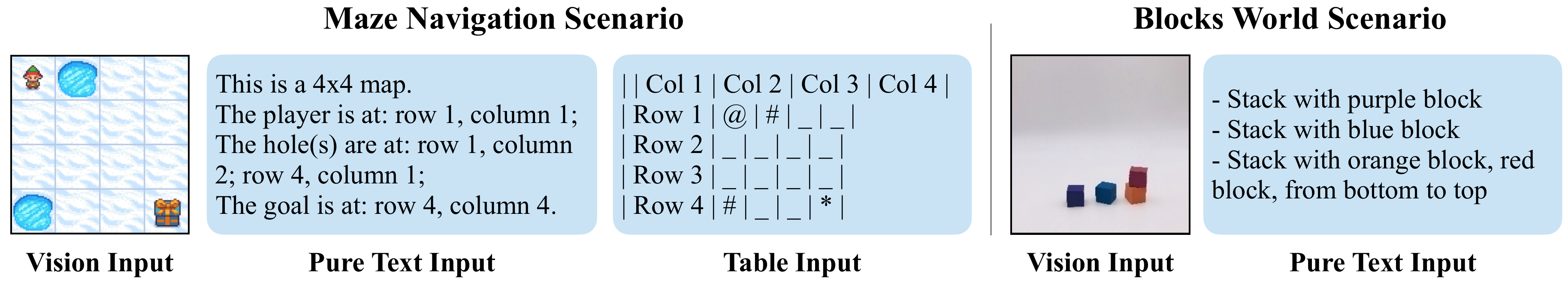}
\end{center}
\caption{The visual and corresponding textual inputs.}
\label{fig:textual-input}
\end{figure*}
\begin{table}[h!]
\small
\centering
\caption{{Decomposed Capability Analysis. Similar to the spatial planning task, each task consists of test with different difficulties. Results better than $70\%$ are \textbf{bolded}. Please refer to Appendix~\ref{table-different-level} for the complete evaluation results for different difficulties.}}    \label{tab:decomposed-zero}
\vskip 1ex
\resizebox{0.75\textwidth}{!}{
\begin{tabular}{@{}lcccccccc@{}}
\toprule
     & \multicolumn{4}{c}{\textsc{Maze Navigation}}
       & \multicolumn{4}{c}{\textsc{Blocksword}} \\
     \cmidrule(lr){2-5}
     \cmidrule(lr){6-9}
     Task & {\small{\textbf{T1}}} & {\small{\textbf{T2}}} & {\small{\textbf{T3}}} & {\small{\textbf{T4}}} & {\small{\textbf{T1}}} & {\small{\textbf{T2}}} & {\small{\textbf{T3}}} & {\small{\textbf{T4}}} \\
     \midrule
{\small Random Guess     } &  0.5 & 0.25 & 0.25             & 0.5 & 0.17 & 0.25 & 0.25 & 0.5          \\
\midrule
{\small Gemini \cite{team2023gemini}} &   0.58 &	0.56	&0.33&	0.49&\textbf{0.86} & 0.51 &	0.54 &	0.55  \\
{\small GPT-Vision \cite{achiam2023gpt}  } & 0.56  & 0.27 & 0.46	& 0.56	&	\textbf{0.73} &	\textbf{0.80} & \textbf{0.70} &	\textbf{0.71}  \\ 
{\small Claude-3 \cite{anthropic2024claude}   } &   0.45 & 0.67 & 0.32 & 0.61 & 0.43 & 0.53 & 0.49 & 0.66          \\ 
{\small GPT-4o \cite{GPT-4o}  } & 0.58 & 0.67 & 0.58 & \textbf{0.74} & \textbf{0.95} & \textbf{0.90} & \textbf{0.90} & \textbf{0.76} \\ 
\midrule
{\small LLaVA \cite{liu2024visual}     } &  0.49 & 0.27 & 0.21 & 0.54 & 0.22 & 0.21 & 0.24 & 0.55  \\
{\small InternLM \cite{dong2024internlm}  } &  0.48 &	0.27&	0.29&	0.58&	0.25&	0.32&  0.26 & 	0.53  \\
{\small InternLM-VL \cite{dong2024internlm}                   } &  0.41	&0.20	&0.17&	0.47	&0.22	&0.20 & 0.20 & 0.53  \\ 
{\small InstructBLIP \cite{dai2024instructblip}                  }  & 0.44 &	0.23 & 	0.21 &	0.37 & 0.21 & 0.16 & 0.22 & 0.47\\
{\small SPHINX \cite{lin2023sphinx}                   } &   0.56 &	0.28 &	0.32 &	0.59 &	0.24 &	0.33 & 0.27 & 0.58 \\ \bottomrule
\end{tabular}
\vspace{-2ex}
}

\end{table}

\subsection{The Perception and Reasoning Sub-tasks Evaluation} 
From the previous observation, we identify that spatial \textit{perception} and \textit{reasoning} could be two important capabilities for an agent to successfully perform visual spatial planning.
Next, we evaluate the perception and reasoning abilities through the remaining tasks T1-T4. Similar to previous setting, all the evaluation is conducted under zero-shot settings. 

The performance results are presented in Table~\ref{tab:decomposed-zero}. 
We observe that the recent GPT-4o and GPT-Vision achieve good performance across a series of tasks, demonstrating a decent capability in perception and reasoning.
However, the overall performance of private models hovers around $50\%$ accuracy, which is far from satisfactory for agents requiring spatial intelligence.
Furthermore, the performance of open-source models is mostly close to random guessing on these tasks, indicating a significant gap compared to private models. 
Besides, we note that tasks T1-T3 focus on perception abilities, and task T4 involves the ability of both understanding input images and perform reasoning. We perform further analysis to disentangle these two abilities in Section~\ref{visual-vs-text-sec}.

\subsection{The Effects of Visual Input Perception and Reasoning}\label{visual-vs-text-sec}
Previous analysis shows that even current state-of-the-art models have clear deficiencies in various aspects of visual spatial planning. In this study, we focus on 
disentangling the effects of perception and reasoning
by exploring the performance gain assuming the model had perfect perception.

The key strategy of this study is to create a scenario where the model has already acquired all the necessary information that would typically be obtained through visual perception.
To this end, for every input image, we produce the corresponding textual inputs and replace those images, as demonstrated in Figure~\ref{fig:textual-input}. 
For the maze navigation scenario, we use either pure text descriptions or tables to depict the image. For the blocks world scenario, we use pure text descriptions. We do not use tables for the blocks world scenario because the number of blocks in each horizontal stack is usually unequal, making it difficult to form a complete table. Please refer to Appendix~\ref{prompt-textual-input} for a complete example with pure text or table input. 

\begin{figure*}[t!]
\begin{center}
\includegraphics[width=1\textwidth]{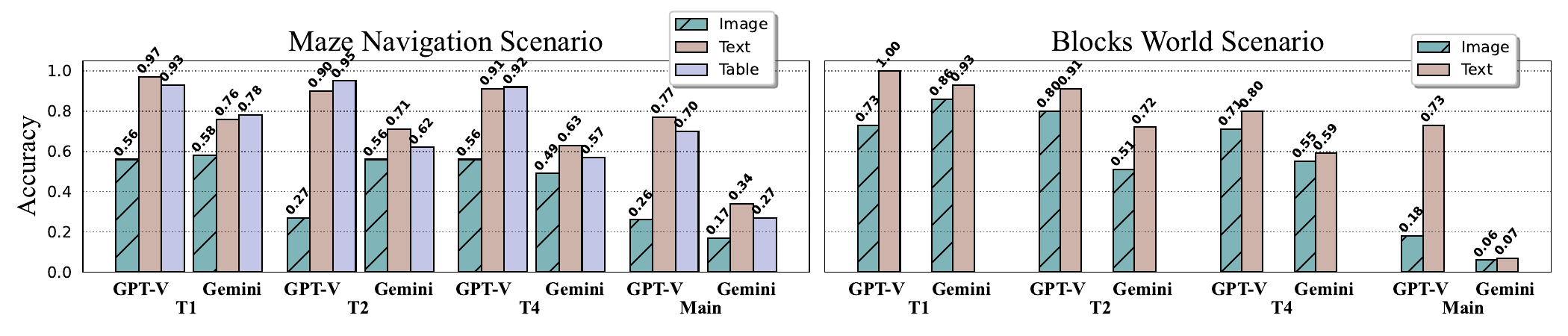}
\end{center}
\vskip -0.2in

\caption{Performance comparison with the visual/textual input. When the environment is described by text instead of image, the performance increases significantly.}
\label{fig:visual-vs-textual}
\end{figure*}

The results are shown in Figure~\ref{fig:visual-vs-textual}. We observe a clear performance improvement when using textual input across every task. This suggests image perception presents significant challenges for VLMs, and poor perception ability is a key factor in the inferior performance observed in previous tasks. 
Meanwhile, we observe that even with textual input, Gemini still cannot achieve decent performance on tasks that require reasoning. This indicates deficiencies in its reasoning capabilities as well.

\begin{table}[!t]
\small
\centering
{ 
    \caption{{Effects of providing in-context examples.}}
\resizebox{0.92\linewidth}{!}{
\begin{tabular}{@{}lcccccccccc@{}}
\toprule
     & \multicolumn{5}{c}{\textsc{Maze Navigation}}
       & \multicolumn{5}{c}{\textsc{Blocksword}} \\
     \cmidrule(lr){2-6}
     \cmidrule(lr){7-11}
     Task & {\small{\textbf{T1}}} & {\small{\textbf{T2}}} &{\small{\textbf{T3}}} & {\small{\textbf{T4}}} & {\small{\textbf{Main}}} & {\small{\textbf{T1}}} & {\small{\textbf{T2}}} &{\small{\textbf{T3}}} & {\small{\textbf{T4}}} & {\small{\textbf{Main}}} \\
     \midrule
{\small Gemini, 0-shot} &   0.58 &	0.56&	0.33& 0.49 & 0.17 & 0.86 &	0.51 &	0.54 & 0.55 & 0.03        \\ 
{\small Gemini, 1-shot     } & 0.50  & 0.66 & 0.31            & 0.48 & 0.20 & 0.91 & 0.68 & 0.71 & 0.59 & 0.03         \\
{\small Gemini, 2-shot     } & 0.53  & 0.68 & 0.31            & 0.51 & 0.21 & 0.90 & 0.76 & 0.70 & 0.61  & 0.03       \\
{\small Gemini, 4-shot     } &  0.53 & 0.67 & 0.35            & 0.53 & 0.19 & 0.91 & 0.64 & 0.69 & 0.62 & 0.06       \\
\midrule
{\small GPT-Vision, 0-shot } & 0.56  & 0.27  & 0.46& 0.56	& 0.26 &	0.73 & 0.80 & 0.70	& 0.71 & 0.10 \\
{\small GPT-Vision, 1-shot } & 0.55   &	0.50 &0.47	&	0.57&	0.28 & 0.89 & 0.84& 0.94 & 0.73 & 0.11 \\
{\small GPT-Vision, 2-shot } & 0.55   & 0.63	 &	0.50&	0.56& 0.30	& 0.90 & 0.83& 0.95 & 0.71 & 0.16  \\
{\small GPT-Vision, 4-shot } & 0.54   &	0.69 &	0.54&	0.56& 0.29	& 0.90 & 0.79& 0.96 & 0.73 & -  \\
\bottomrule
\end{tabular}
}
    \label{tab:in-context}
}
\vskip -0.1in
\end{table}
\begin{table}[!t]
\small
\centering
{ 
    \caption{{Fine-tuning results for open-source models.}}
    \resizebox{0.92\linewidth}{!}{
\begin{tabular}{@{}lccccccccccc@{}}
\toprule
     & & \multicolumn{5}{c}{\textsc{Maze Navigation}}
       & \multicolumn{5}{c}{\textsc{Blocks of World}} \\
     \cmidrule(lr){3-7}
     \cmidrule(lr){8-12}
     Model & Setting & {\small{\textbf{T1}}} & {\small{\textbf{T2}}} &{\small{\textbf{T3}}} & {\small{\textbf{T4}}} & {\small{\textbf{Main}}} & {\small{\textbf{T1}}} & {\small{\textbf{T2}}} &{\small{\textbf{T3}}} & {\small{\textbf{T4}}} & {\small{\textbf{Main}}} \\
     \midrule
\multirow{2}{*}{LLaVA} & zero-shot &   0.49 & 0.27 & 0.21 & 0.54 & 0.05 & 0.22 & 0.21 & 0.24 & 0.55 &  0.01      \\ 
 & fine-tune   & 0.53 &	0.99 &	0.51 & 0.93 & 0.60 & 1.00 & 1.00 &	1.00 &	1.00 & 0.97         \\ 
\multirow{2}{*}{InternLM} & zero-shot & 0.48 &	0.27&	0.29&	0.58&0.11	& 0.25&	0.32&  0.26 & 	0.53 & 0.00         \\
 & fine-tune & 0.52  & 0.59 & 0.91 & 0.59 & 0.17            & 0.29 & 0.44 & 0.69 & 0.62 & 0.09  \\
\bottomrule
\end{tabular}
}
    \label{tab:training}
}
\vskip -0.1in
\end{table}

\subsection{In-context Learning in Visual Spatial Planning}
\label{subsec:icl}
In-context learning is a widely-adopted method to enhance LM's reasoning ability~\cite{brown2020language}. In this analysis, we study if it boosts the visual spatial planning capabilities. We included varying numbers of examples for Gemini and GPT-Vision (refer to Appendix~\ref{prompt-ic-example} for the input examples). The result is shown in Table~\ref{tab:in-context}. There are two key observations: First, in-context examples make some potential contributions, but they are not significant. Introducing examples only benefits in several sparse cases, such as \textbf{T2} in maze navigation and \textbf{T3} in blocksworld. Second, scaling in-context examples generally does not help, as illustrated by the saturated performance in each task. 

\subsection{Fine-tuning in VSP Tasks}
\label{subsec:finetuning}
Finally, we assess the capabilities of the open-source model through dedicated training for each task. In these tasks, the model is trained on 10k data points. We use the default hyperparameters provided in the official repo. The results, shown in Table~\ref{tab:training}, demonstrate clear performance improvements for both models across a series of tasks, highlighting their potential in spatial planning. Additionally, we observe that LLaVA shows greater improvement compared to InternLM, suggesting that different model architectures may exhibit varying levels of efficacy in spatial planning capabilities.

\section{Conclusion}
We present VSP, a benchmark measuring and diagnosing the visual spatial planning capabilities in VLMs. The VSP quantifies the model’s performance through a series of carefully designed tasks, with main tasks focusing on the general spatial planning abilities and sub-tasks focusing on the individual capabilities needed for the main task.  Experiments on current models show that both private models and open-source models fail to generate effective plans for even simple spatial planning tasks, and further analyses expose their bottlenecks in spatial perception and reasoning abilities. Our work illuminates future directions for improving VLMs' abilities in spatial planning.

\bibliography{neurips-2}
\bibliographystyle{unsrtnat}

\newpage
\clearpage
\appendix
\section{Details of Benchmarks}\label{detail-task}
\subsection{Additional Task Implementation Process and Statistics}
In Section~\ref{main-text-benchmark-creation}, we described our general benchmark creation process. Our task creation process can be divided into three stages:
(a) preparing the input image;
(b) formulating task prompts;
(c) developing scripts for auto evaluation.
Additionally, some specific tasks require extra steps to implement, such as question and answer generation. In what follows, we describe these steps in detail.

\textbf{Maze Navigation, Main Task} In this task, agents need to find a safe path to navigate from the start grid to the goal. We adjust the map generation mechanisms to ensure the positions of the start grid, the goal, and the holes are all randomly generated, while ensuring there is at least one viable safe path from the start grid to the goal. For each grid, the probability that it contains the hole is 20\%.

\textbf{Maze Navigation, T1} In this task, agents need to identify whether a specific grid is safe (i.e., whether it contains a hole or not). We randomly sample a row number and a column number and ask the safety question for this randomly chosen grid in each problem of this task. Additionally, to prevent the model from patterned guessing and achieving falsely high ratings (e.g., answering "not safe" for all images and obtaining high accuracy scores), we regenerate the map for this task to ensure that the safe and unsafe grids each comprise around 50\% of the total grids in a single map.

\textbf{Maze Navigation, T3} In this task, agents need to find the correct textual description that fits the visual environment. For each problem, we prepare four textual description candidates. One candidate is the correct answer, one candidate has the correct size but an incorrect map arrangement, and the other two candidates have the wrong size. The candidates are shuffled to prevent the model from making random guesses.

\textbf{Maze Navigation, T4} In this task, agents need to determine if the given action series is safe or not. Similarly, to prevent the models from achieving falsely high ratings through guessing, we generate the action series to ensure that around 50\% of them are safe and the other 50\% are not. For the unsafe paths, the particular step in which the player steps into a hole is also randomly chosen.

\textbf{Blocks World, T2} In this task, agents need to determine the spatial relation between two designated blocks. In addition to the directional relation (``above'' and ``below''), we note that it is important for agents to recognize if two blocks are at the same stack or not. Therefore, we design the following four candidates for this question: (A) The first block is directly above the second block, and they are in the same stack; (B) The first block is directly below the second block, and they are in the same stack; (C) The two blocks are at different stacks; (D) At least one of the mentioned blocks do not exist in the presented image.

\textbf{Blocks World, T4} In this task, agents need to determine the consequence of a given moving plan. Specifically, some invalid moving plans contain actions that cannot be executed in the given scenario, such as trying to move a block that is covered by another block. To prevent guessing in this task, similar to the maze navigation scenario, we generate the action plans to ensure that half of the plan candidates are executable. For the plans that cannot be executed, there can be two types of errors: first, the plan may include steps that involve moving a block from or to an invalid position; second, the plan may try to move a block that does not exist. We randomly generate these errors in inputs.

\textbf{Statistics} The VSP benchmark consists of 10 tasks in two scenarios. For each task, the problems are designed with different difficulty levels. Specifically, each difficulty level consists of 100 problems. In total, the VSP benchmark includes 4.4k questions.

\subsection{Complete Prompt}
In this subsection, we provide the complete prompts for each task. Generally, the prompts consist of a general task description at the beginning and a specific question at the end. The prompts interleave text and images in a pattern similar to a human-readable manual with reference figures.

\begin{tcolorbox}[left=1.2pt,right=1.2pt,top=1.2pt,bottom=1.2pt]
\small
\textbf{Prompt for Maze Navigation scenario, Main task (Spatial Planning):}
\\
\\
As a professional maze solver, your task is to analyze a grid-based map and devise an action plan that enables a player to reach the goal from the starting point without falling into any holes, using the fewest possible moves. Since coding is not within your skill set, your approach relies on logical reasoning of the map.
\\
\\
\#\# Game Setup\\
- The game presents a fully observable grid-based map.\\
- The player starts at a specified grid square, with the goal located elsewhere on the map.\\
- Each grid square is either safe or contains a hole.\\
- Your goal is to guide the player to the goal while avoiding holes.\\
The following figure shows how the player, the holes (non-safe grid), the lands (safe grids), and the goals look like.\\
\\
\includegraphics[width=0.3\textwidth]{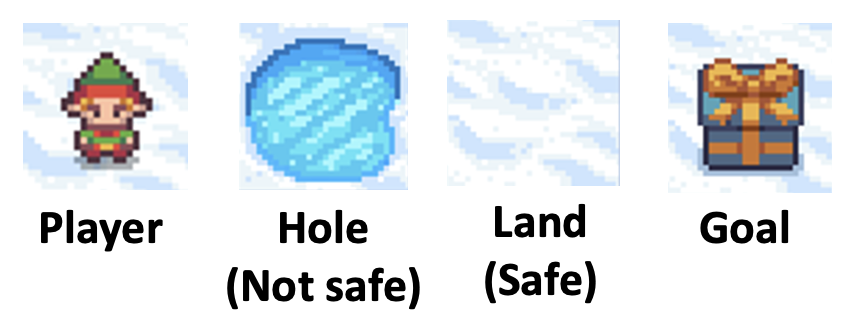}
\\
\\
\#\# Moving Rules\\
- The action plan involves a series of moves: `L' (left), `R' (right), `U' (up), or `D' (down).\\
- Each move transfers the player to the adjacent square in that direction, provided it is a safe square. The player cannot move more than one square at a time.\\
- Moving off the edge of the map has no effect. The player will remain at the same square.\\
- DO NOT MOVE INTO A HOLE! Falling into a hole results in defeat.\\
- Locating at the grid containing the goal results in victory.\\
We provide an example to further illustrate the rules.\\
\\
\includegraphics[width=0.25\textwidth]{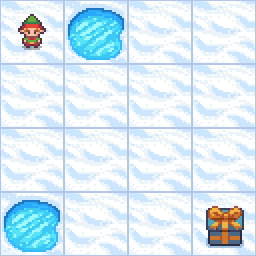}
\\
\\
In this provided example:\\
- The player is at Row 1, Column 1;\\
- The goal is at Row 4, Column 4;\\
- There are two holes: one at Row 1, Column 2, and another at Row 4, Column 1.\\
- The player can move DOWN. This is because moving down brings them to Row 2, Column 1, and this cell is safe (without holes).\\
- Moving UP has no effects. This is because the player is already in the topmost row.\\
- Similarly, moving LEFT has no effects because the player is already in the left-most column.\\
- Moving RIGHT places the player at Row 1, Column 2. Since there is a hole at this grid, this move results in a loss.\\
\\
\#\# Procedure and Output\\
Now you will solve the given maze. To solve it, please generate text exactly follow the following steps:\\
1. First, interpret map. List where the player is at now, where is the goal, and where are the holes.\\
2. Then, generate an action plan to navigate to the goal step by step. At each step, you should check:\\
(a) Where the current move leads the player to (the row and column);\\
(b) What is in that grid. Is it a hole? Is it the goal? Is it an empty space?\\
(c) Determine if that is a safe action. If not, correct it and re-generate the action plan.\\
3. Next, verify if the steps successfully navigate the player to the goal without falling into the hole. If not, restart from step 2 and re-generate this step.\\
\textit{(continue in next page)}
\end{tcolorbox}

\begin{tcolorbox}[left=1.2pt,right=1.2pt,top=1.2pt,bottom=1.2pt]
\small
\textit{(Continued)}
\\
4. If succeed, output an aggregated plan using "Action plan: <PLAN>", where <PLAN> is a string concatenated action in each step. For example, "Action plan: L,L,R,U,D" meaning an action plan of left, left, right, up, and down. Double check the final action plan is consistent with the previous analysis.\\
Do not output any extra content after the above aggregated output.\\
\\
Please generate the action plan for the following maze:
\\
\\
\includegraphics[width=0.25\textwidth]{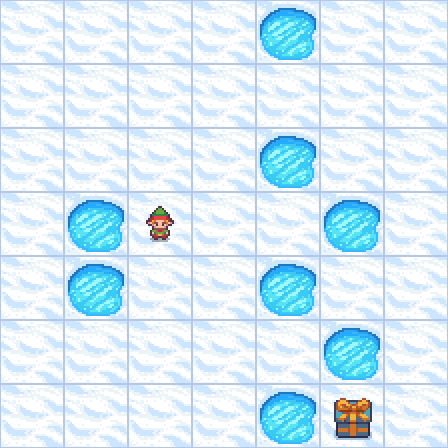}
\end{tcolorbox}

\begin{tcolorbox}[left=1.2pt,right=1.2pt,top=1.2pt,bottom=1.2pt]
\small
\textbf{Prompt for Maze Navigation scenario, Task 1 (Single Object Perception):}
\\
\\
In this task, you will analyze a maze to determine if there is a hole in a specific position.\\
The following figure illustrates the appearances of the player, holes, lands, and the goal within the maze. You will need to focus on the appearance of the hole.
\\
\\
\includegraphics[width=0.3\textwidth]{figures/supp/system-figure-1.png}
\\
\\
Here is an example to illustrate how to analyze and answer the question:
\\
\\
\includegraphics[width=0.25\textwidth]{figures/supp/system-figure-2.png}
\\
Example question: Is there a hole in row 3, column 3?\\
\\
In this example:\\
- We check the position in row 3, column 3.\\
- According to the image, it is a land square. It does not contain a hole.\\
- Therefore, you will output "<Output> No".\\
\\
Your output should be: ``<Output> No'' or ``<Output> Yes'', depending on whether there is a hole at the specified position.
\\
\\
Now you will analyze the following maze and answer the question: 
\\
Is there a hole in row 2, column 1?
\\
\includegraphics[width=0.25\textwidth]{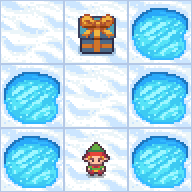}

\end{tcolorbox}

\begin{tcolorbox}[left=1.2pt,right=1.2pt,top=1.2pt,bottom=1.2pt]
\small
\textbf{Prompt for Maze Navigation scenario, Task 2 (Spatial Relation Perception):}
\\
\\
In this task, you will analyze a maze to determine the relative positions of the player and the goal.\\
The following figure illustrates the appearances of the player, holes, lands, and the goal within the maze. You will need to focus on the player and the goal.
\\
\\
\includegraphics[width=0.3\textwidth]{figures/supp/system-figure-1.png}
\\
\\
To describe their relative positions, use the directional indicators from {``Above'', ``Below'', ``Left'', ``Right''}. We provide an example to illustrate how to interpret and describe these positions:
\\
\\
\includegraphics[width=0.25\textwidth]{figures/supp/system-figure-2.png}
\\
In this example:\\
- We focus on the position of the player and the goal.\\
- Rows: The player is at row 1, and the goal is at row 4. Here, the row number is from top to bottom. Comparing player (row=1) with goal (row=4), player is counted first. Therefore, the player is positioned above the target.\\
- Columns: The player is at column 1, and the goal is at column 4. Here, the column number is from left to right. Comparing player (column=1) with goal (column=4). Therefore, the player is to the left of the target.\\
- Remember that we should answer the player's position with respect to the goal, not the opposite. Therefore, we answer ``Above,Left''.
\\
\\
Your output should be two parts:\\
1. Analyze the rows and columns of the player and the goal like shown above. \\
2. Following your analysis, output answer as ``<Output> <Position>''. For example, ``<Output> Above,Left'' means the player is above and to the left of the goal, and ``<Output> Below'' means the player is below the goal. 
Note that you should not output ``Left'' or ``Right'' if the player and the goal are at the same column, and similarly, you should not output ``Above'' or ``Below'' if the player and the goal are at the same row.
\\
\\
Now you will analyze the following maze and determine the relative position of the player in relation to the goal.
\\
\includegraphics[width=0.25\textwidth]{figures/supp/maze-t5-eg.png}

\end{tcolorbox}

\begin{tcolorbox}[left=1.2pt,right=1.2pt,top=1.2pt,bottom=1.2pt]
\small
\textbf{Prompt for Maze Navigation scenario, Task 3 (Environment Perception):}
\\
\\
In this task, you will analyze a maze presented in an image. Later, you will be presented with four choices, each offering a textual representation of a candidate maze. You will need to choose the representation that exactly reflects the contents of the given image.\\
The following figure illustrates the appearances of the player, holes, lands, and the goal within the maze in the image.
\\
\\
\includegraphics[width=0.3\textwidth]{figures/supp/system-figure-1.png}
\\
\\
This is how the player, the holes (non-safe grid), the lands (safe grids), and the goals look like in a map:\\
- The player is represented as "@"\\
- The hole is represented as "\#"\\
- The safe grid is represented as "\_"\\
- The goal is represented as "*"\\
- If the player is at the goal (at this case the game is solved), that grid is represented as "\%"\\
\\
We provide an example to illustrate how to interpret the input, candidates, and answer the question. Here is the image input:
\\
\\
\includegraphics[width=0.25\textwidth]{figures/supp/system-figure-2.png}
\\
Here are the textual candidates:
\\
(A)\\
\texttt{
| | Col 1 | Col 2 | Col 3 |\\
| Row 1 | \# | \_ | \_ |\\
| Row 2 | \# | @ | \# |\\
| Row 3 | \_ | * | \_ |\\
\\
}
(B)\\
\texttt{
| | Col 1 | Col 2 | Col 3 | Col 4 | Col 5 |\\
| Row 1 | \_ | \_ | \_ | \_ | \_ |\\
| Row 2 | \_ | \# | \_ | \_ | \_ |\\
| Row 3 | \_ | \# | * | \_ | \# |\\
| Row 4 | \_ | @ | \_ | \_ | \_ |\\
| Row 5 | \_ | \_ | \_ | \# | \_ |\\
\\
}
(C)\\
\texttt{
| | Col 1 | Col 2 | Col 3 | Col 4 |\\
| Row 1 | @ | \# | \_ | \_ |\\
| Row 2 | \_ | \_ | \_ | \_ |\\
| Row 3 | \_ | \_ | \_ | \_ |\\
| Row 4 | \# | \_ | \_ | * |\\
\\
}
(D)\\
\texttt{
| | Col 1 | Col 2 | Col 3 | Col 4 |\\
| Row 1 | \_ | \_ | \_ | \_ |\\
| Row 2 | * | \_ | \_ | \_ |\\
| Row 3 | @ | \_ | \# | \_ |\\
| Row 4 | \_ | \_ | \_ | \# |\\
\\
}
Here is an example of how to analyze and answer the question:\\
\textit{(continue in next page)}
\end{tcolorbox}

\begin{tcolorbox}[left=1.2pt,right=1.2pt,top=1.2pt,bottom=1.2pt]
\small
\textit{(Continued)}\\
- First, we focus on the difference of the maze shape between the candidates and the input image.\\
- We begin by examining the input image. It is a 4-by-4 maze. We then review the candidates. Candidate A is a 3-by-3 maze. Therefore, it is not the correct answer. Similarly, Candidate B is a 5-by-5 maze, which also cannot be correct. Both Candidate C and Candidate D are 4-by-4 mazes. Now we only need to choose from them.\\
- For the remaining candidates, we compare the positions of the players, goals, and the holes in the maze.\\
- We first check the input image. What is the position of the player in the image? The player is in row 1, column 1. We then check the remaining candidates. For Candidate C, the textual representation indicates the player is also at row 1, column 1, matching the input image. For Candidate D, the player is located at row 3, column 1. Hence, Candidate D is not the correct answer.\\
- We double check the remaining Candidate C, and it correctly shows the position of the player, holes, and the goal. It is therefore the correct answer.\\
<Answer> C\\
\\
\\
Your output should consist of two parts:\\
1. First, analysis the input image and candidates similar to the reasoning process above.\\
2. Following the reasoning process, output answer as "<Answer> <Choice>", where "<Choice>" is one of {A,B,C,D}.\\
Important: Note that there will be only one correct answer. If you find no answer or multiple answers, you must go back and recheck your reasoning process. You are not allowed to provide 0 or more than 1 answer.\\
\\
\\
Now answer the question below. Here is the image input:
\\
\includegraphics[width=0.25\textwidth]{figures/supp/maze-t5-eg.png}\\
Here are the textual candidates:\\
\\
<CANDIDATES>
\end{tcolorbox}

\begin{tcolorbox}[left=1.2pt,right=1.2pt,top=1.2pt,bottom=1.2pt]
\small
\textbf{Prompt for Maze Navigation scenario, Task 4 (Spatial Reasoning):}\\
\\
You are a maze-solving agent playing a pixelated maze video game.\\
Mazes are presented on grid maps, where each tile can be empty land, or contain a player, hole, or goal.\\
Each of the above tile types are represented as square pixel art images.\\
\\
In this task, you will analyze a grid-based map and determine if a provided action plan is safe. A safe action plan avoids stepping into holes in the map.\\
The following figure illustrates the appearances of the player, holes, lands, and the goal within the maze.\\
\\
\includegraphics[width=0.3\textwidth]{figures/supp/system-figure-1.png}
\\
\\
\#\# Moving Rules\\
- The action plan involves a series of moves: 'L' (left), 'R' (right), 'U' (up), or 'D' (down).\\
- Each move transfers the player to the adjacent square in that direction, provided it is a safe square. The player cannot move more than one square at a time.\\
- Moving off the edge of the map has no effect. The player will remain at the same square.\\
- DO NOT MOVE INTO A HOLE! Falling into a hole results in defeat.\\
- Locating at the grid containing the goal results in victory.\\
We provide an example to further illustrate the rules.\\
\textit{(continue in next page)}
\end{tcolorbox}

\begin{tcolorbox}[left=1.2pt,right=1.2pt,top=1.2pt,bottom=1.2pt]
\small
\textit((Continue))
\\
\includegraphics[width=0.25\textwidth]{figures/supp/system-figure-2.png}
\\
In this provided example:\\
- The player is at Row 1, Column 1;\\
- The goal is at Row 4, Column 4;\\
- There are two holes: one at Row 1, Column 2, and another at Row 4, Column 1.\\
- The player can move DOWN. This is because moving down brings them to Row 2, Column 1, and this cell is safe (without holes).\\
- Moving UP has no effects. This is because the player is already in the topmost row.\\
- Similarly, moving LEFT has no effects because the player is already in the left-most column.\\
- Moving RIGHT places the player at Row 1, Column 2. Since there is a hole at this grid, this move results in a loss.\\
\\
\#\# Procedure and Output\\
Your output should include the following parts:\\
1. First, interpret map. List where the player is at now, where is the goal, and where are the holes.\\
2. Then, reasoning by following the given action plan. At each step, you should check:\\
(a) Where the current move leads the player to (the row and column);\\
(b) What is in that grid. Is it a hole? Is it the goal? Is it an empty space?\\
(c) Determine if that is a safe action.\\
3. Output if the action sequence is safe using "<Output> Yes" or "<Output> No". A safe action sequence should not include any unsafe actions.\\
\\
Now please determine if the action sequence is safe for this given maze:\\
\includegraphics[width=0.25\textwidth]{figures/supp/maze-t5-eg.png}\\
The action sequence is:\\
Up, Down, Right, Down, Right
\end{tcolorbox}

\begin{tcolorbox}[left=1.2pt,right=1.2pt,top=1.2pt,bottom=1.2pt]
\small
\textbf{Prompt for Block World scenario, Main Task (Spatial  Planning):}\\
\\
You are a robot that sorts and organizes colored blocks by adding and removing them to stacks.\\
You can move them between stacks to produce a desired end state.\\
\\
In this task, you will see two photos of blocks. These photos show the beginning and end state of the blocks. Your task is to find a shortest movement plan to transit from the beginning state to the end state. Since coding is not within your skill set, your approach relies on logical reasoning of the map.\\
\\
\#\# Game Setup\\
- The stacks of blocks are presented in images. You must view and interpret the image in order to determine which blocks are in which stack and determine how to move them.\\
- Each block has a unique color (blue, yellow, purple, orange, red, green).\\
- Blocks are stacked vertically in a stack, forming multiple stacks. All stacks are on the table.\\
- In a single move, you can only move the top block of any pile. Attempting to move lower blocks is considered an invalid move.\\
- You can either (a) move the top block to the top of another stack, or (b) place the top block on the table, creating a new stack with just one block.\\
\\
We provide an example to further illustrate the rules:\\
\includegraphics[width=0.2\textwidth]{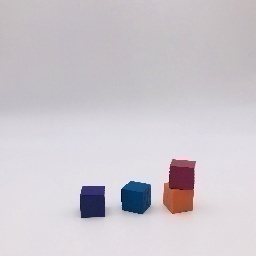}\\
This example features four blocks arranged in three stacks:\\
- Stack 1: Purple block (alone)\\
- Stack 2: Blue block (alone)\\
- Stack 3: From bottom to top: Orange block, Red block\\
You can only move the top block of each stack: the purple block, the blue block, and the red block. The orange block is stuck underneath the red block and cannot be moved directly.\\
Each move can place the block on another stack or on the table (creating a new stack of one).  For instance, you could move the red block to either the blue stack or the table.\\
**Important Note**: The order of the stacks doesn't matter in this game. Two images are considered equivalent as long as the stacks contain the same blocks, regardless of the order in which the stacks appear. For example, an image with stack A on the left and stack B on the right is equivalent to an image with stack B on the left and stack A on the right.\\
\\
\#\# Procedure and Output\\
Your output should follow this format:\\
1. First, analyze the starting and ending configurations, including the number of stacks and the blocks in each stack (similar to the example above).\\
2. Then, list the moves in a step-by-step manner using the format move(SOURCE, TARGET). Remember, "SOURCE" refers to the block being moved (always the top block of a stack), and "TARGET" refers to the destination (another stack or the table).

\#\# Example Output\\
<Analysis> \\
Starting state: there are three stacks:\\
- Stack 1: Purple block (alone)\\
- Stack 2: Blue block (alone)\\
- Stack 3: From bottom to top: Orange block, Red block\\
Ending state: there are three stacks:\\
- Stack 1: Purple block (alone)\\
- Stack 2: From bottom to top: Orange block, Blue block\\
- Stack 3: Red block (alone)\\
<Output>\\
1. move(red,table)\\
2. move(blue,orange)\\

\textit{(continue in next page)}
\end{tcolorbox}

\begin{tcolorbox}[left=1.2pt,right=1.2pt,top=1.2pt,bottom=1.2pt]
\small
\textit{(Continue)}\\
Now please generate moving plan. The beginning state is:\\
\includegraphics[width=0.2\textwidth]{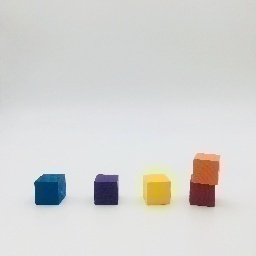}
\\
The end state is:\\
\includegraphics[width=0.2\textwidth]{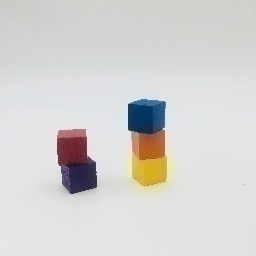}
\end{tcolorbox}

\begin{tcolorbox}[left=1.2pt,right=1.2pt,top=1.2pt,bottom=1.2pt]
\small
\textbf{Prompt for Block World scenario, Task 1 (Single Object Perception):}\\
\\
In this task, you will see a photo of blocks. You will analyze the block configuration and then answer a question regarding the color of blocks in a specific place.\\
\\
\#\# Game Setup\\
- Each block has a unique color (blue, yellow, purple, orange, red, green).\\
- Blocks are stacked vertically in a stack, forming multiple stacks.\\
- In the questions, the position of the blocks is represented as ``Stack s, Level l''. The stack number is counted from left to right, and the level number is counted from bottom to top.\\
We provide an example to further illustrate the setting:
\\
\includegraphics[width=0.2\textwidth]{figures/supp/example0.jpg}
\\
In this example, there are four blocks in three stacks. From left to right:\\
- Stack 1 has one level. Level 1 contains a purple block.\\
- Stack 2 has one level. Level 1 contains a blue block.\\
- Stack 3 has one level. From bottom to top: level 1 has an orange block, and level 2 has a red block.\\
As such, for the question ``What is the color of the block at stack 3, level 1?", the correct answer is "<Output> orange''.\\
\\
\#\# Procedure and Output\\
Your output should follow this format: \\
1. First, analyze the block configuration;\\
2. Then, answer the question with the format <Output> <Color>, where <Color> is one of (blue, yellow, purple, orange, red, green). For example, ``<Output> red''.\\
\\
Now please answer the following question based on the given image below:\\
What is the color of the block at stack 1, level 2?
\\
\includegraphics[width=0.2\textwidth]{figures/supp/0-output.jpg}
\end{tcolorbox}

\begin{tcolorbox}[left=1.2pt,right=1.2pt,top=1.2pt,bottom=1.2pt]
\small
\textbf{Prompt for Block World scenario, Task 2 (Spatial Relation Perception):}\\
\\
In this task, you will see a photo of blocks. You will analyze the block configuration and then answer a question regarding the spatial relation of two specified blocks.\\
\\
\#\# Game Setup\\
- Each block has a unique color (blue, yellow, purple, orange, red, green).\\
- Blocks are stacked vertically in a stack, forming multiple stacks.\\
- The possible relations of two blocks include: (A) One block is directly above another block, and they are at the same stack; (B) One block is directly below another block, and they are at the same stack; (C) Two blocks are at different blocks; (D) At least one of the asked blocks does not exist in the image.\\
\\
We provide an example to further illustrate the rules:\\
\includegraphics[width=0.2\textwidth]{figures/supp/example0.jpg}
\\
In this example, there are four blocks in three stacks. From left to right:\\
- Stack 1: Purple block (alone)\\
- Stack 2: Blue block (alone)\\
- Stack 3: From bottom to top: Orange block, Red block\\
We can answer the question regarding the spatial relations accordingly. For example, for the question ``What is the spatial relation between the purple block and blue block?'', since the purple block and the blue block are at different stacks, we should choose the choice indicating they are at different stacks.\\
\\
\#\# Procedure and Output\\
Your output should follow this format: \\
1. First, analyze the block configuration;\\
2. Then, answer the question with the format <Output> <Choice>, where <Choice> is one of {A,B,C,D}. For example, ``<Output> A''.\\
\\
Now please answer the following question based on the given image below:\\
What is the spatial relation between the red block and the yellow block?\\
(A) The red block is directly above the yellow block, and they are in the same stack.\\
(B) The red block is directly below the yellow block, and they are in the same stack.\\
(C) The red block and the yellow block are at different stacks.\\
(D) At least one of the red block and the yellow block does not exist in the presented configurations.\\

\includegraphics[width=0.2\textwidth]{figures/supp/0-output.jpg}

\end{tcolorbox}

\begin{tcolorbox}[left=1.2pt,right=1.2pt,top=1.2pt,bottom=1.2pt]
\small
\textbf{Prompt for Block World scenario, Task 3 (Environment Perception):}\\
\\
In this task, you will analyze an image containing several stacks of blocks. Later, you will be presented with four choices, each offering a textual representation of a block configuration. You will need to choose the configuration that exactly reflects the contents of the given image.\\
\#\# Game Setup\\
- Each block has a unique color (blue, yellow, purple, orange, red, green).\\
- Blocks are stacked vertically in a stack, forming multiple stacks.\\
This is an image input example:\\
\includegraphics[width=0.2\textwidth]{figures/supp/example0.jpg}

This example features four blocks arranged in three stacks:\\
- Stack 1: Purple block (alone)\\
- Stack 2: Blue block (alone)\\
- Stack 3: From bottom to top: Orange block, Red block\\
Here are examples of textual representations:\\
(A)\\
- Stack with red block, yellow block, from bottom to top\\
- Stack with orange block, purple block, green block, from bottom to top\\
\\
(B)\\
- Stack with purple block\\
- Stack with blue block\\
- Stack with orange block, red block, from bottom to top\\
\\
(C)\\
- Stack with orange block\\
- Stack with purple block\\
- Stack with blue block\\
- Stack with green block, yellow block, from bottom to top\\
\\
(D)\\
- Stack with green block\\
- Stack with yellow block, blue block, from bottom to top\\
- Stack with red block, orange block, from bottom to top\\
\\
We can analyze which text representation exactly reflects the configurations in the image accordingly. In this example:\\
- The input image has 3 stacks, while Candidate A only has 2 stacks. Therefore, Candidate A is not the correct answer.\\
- Similarly, Candidate C has 4 stacks, which also cannot be correct.\\
- For Candidate B, the blocks in each stack match what's shown in the image. This is the correct answer.\\
- For Candidate D, the blocks in each stack do not match the image. For example, stack 1 in the image has a purple block, and there is no any purple block in Candidate D. So this is incorrect.\\
- Therefore, the final answer is B.\\
\#\# Procedure and Output\\
Your output should follow this format:\\
1. First, analyze the block configuration in the image and candidates as shown above;\\
2. Then, answer the question with the format <Output> <Choice>, where <Choice> is one of {A,B,C,D}. For example, "<Output> A".\\
\\
Now please choose the correct textual representation based on the given image below:\\
\includegraphics[width=0.2\textwidth]{figures/supp/0-output.jpg}
\\
Here are the textual candidates:\\
<CANDIDATES>
\end{tcolorbox}

\begin{tcolorbox}[left=1.2pt,right=1.2pt,top=1.2pt,bottom=1.2pt]
\small
\textbf{Prompt for Maze Navigation scenario, Task 4 (Spatial Reasoning):}\\
\\
You are a robot that sorts and organizes colored blocks by adding and removing them to stacks. You can move them between stacks to produce a desired end state.\\
In this task, you will see a photo of blocks. This photo shows the beginning state of the blocks. You will see a photo of blocks. This photo shows the beginning state of the blocks. Meanwhile, you will be provided an action sequence about moving blocks. Your task is to determine if the provided action plan can be successfully executed.
\\
\#\# Game Setup\\
- The block configuration is presented in the image. You must view and interpret the image in order to determine which blocks are in which stack and determine the consequence of moving.\\
- Each block has a unique color (blue, yellow, purple, orange, red, green).\\
- Blocks are stacked vertically in a stack, forming multiple stacks.\\
- A valid action can only move the top block of any stacks. Attempting to move lower blocks is considered an invalid move.\\
- For the destination, a valid move can either (a) move the top block to the top of another stack, or (b) place the top block on the table, creating a new stack with just one block.\\
We provide an example to further illustrate the rules:\\
\includegraphics[width=0.2\textwidth]{figures/supp/example0.jpg}
\\
The sequence of actions provided is: 1. move(red,table) 2. move(green,table)\\
In this example, there are four blocks in three stacks. The stacks are:\\
- Stack 1: Purple block (alone)\\
- Stack 2: Blue block (alone)\\
- Stack 3: From bottom to top: Orange block, Red block\\
It is valid to move the purple block, the blue block, and the red block, since they are at the top of a stack. It is invalid to move the orange block since it is not at the top of a stack (because it is covered by the red block).\\
Each move can place the block on top of another stack or on the table (creating a new stack of one).  For instance, you could move the red block to either the blue stack or the table.\\
\#\# Procedure and Output\\
Your output should follow this format:\\
1. First, briefly analyze the block configuration, and check each action step by step to see if the provided step is valid as shown above.\\
2. Then, answer the question with the format "<Output> Yes" or "<Output> No" to indicate if the action sequence is valid.\\
Here is an example for the output:\\
<Analysis> In the image, there are three stacks:\\
- Stack 1: Purple block (alone)\\
- Stack 2: Blue block (alone)\\
- Stack 3: From bottom to top: Orange block, Red block\\
The first action ``move(red,table)'' is valid, because the red block is on top of a stack (stack 3 in this case), and the target is ``table''. After the first action, the state will become:\\
- Stack 1: Purple block (alone)\\
- Stack 2: Blue block (alone)\\
- Stack 3: Orange block (alone)\\
- Stack 4: Red block (alone)\\
The second action ``move(green,table)'' is invalid, because there is no green block.\\
Therefore, the provided action sequence is invalid.\\
<Output> No\\
\\
Now please determine if the provided action sequence is valid given the following input state:
\\
\includegraphics[width=0.2\textwidth]{figures/supp/0-output.jpg}
\\
The action sequence is: 1. move(red,table) 2. move(yellow,red) 3. move(purple,table)
\end{tcolorbox}

\section{Prompt for textual input}\label{prompt-textual-input}
In Section~\ref{visual-vs-text-sec}, we described the procedure of using textual representation instead of visual input. Below, we use the main task in the maze navigation scenario as an example to show the complete prompt after making the replacement.
\begin{tcolorbox}[left=1.2pt,right=1.2pt,top=1.2pt,bottom=1.2pt]
\small
\textbf{Prompt for Maze Navigation scenario, Main task (Spatial Planning, Textual Input):}
\\
\\
As a professional maze solver, your task is to analyze a grid-based map and devise an action plan that enables a player to reach the goal from the starting point without falling into any holes, using the fewest possible moves. Since coding is not within your skill set, your approach relies on logical reasoning of the map.
\\
\\
\#\# Game Setup\\
- The game presents a fully observable grid-based map.\\
- The player starts at a specified grid square, with the goal located elsewhere on the map.\\
- Each grid square is either safe or contains a hole.\\
- Your goal is to guide the player to the goal while avoiding holes.\\
\\
\#\# Moving Rules\\
- The action plan involves a series of moves: `L' (left), `R' (right), `U' (up), or `D' (down).\\
- Each move transfers the player to the adjacent square in that direction, provided it is a safe square. The player cannot move more than one square at a time.\\
- Moving off the edge of the map has no effect. The player will remain at the same square.\\
- DO NOT MOVE INTO A HOLE! Falling into a hole results in defeat.\\
- Locating at the grid containing the goal results in victory.\\
\\
We provide an example to further illustrate the rules.\\
Example Input:\\
This is a 4x4 map.\\
The player is at: row 1, column 1;\\
The hole(s) are at: row 1, column 2; row 4, column 1;\\
The goal is at: row 4, column 4.\\
\\
In this provided example:\\
- The player is at Row 1, Column 1;\\
- The goal is at Row 4, Column 4;\\
- There are two holes: one at Row 1, Column 2, and another at Row 4, Column 1.\\
- The player can move DOWN. This is because moving down brings them to Row 2, Column 1, and this cell is safe (without holes).\\
- Moving UP has no effects. This is because the player is already in the topmost row.\\
- Similarly, moving LEFT has no effects because the player is already in the left-most column.\\
- Moving RIGHT places the player at Row 1, Column 2. Since there is a hole at this grid, this move results in a loss.\\
\#\# Procedure and Output\\
Now you will solve the given maze. To solve it, please generate text exactly follow the following steps:\\
1. First, interpret map. List where the player is at now, where is the goal, and where are the holes.\\
2. Then, generate an action plan to navigate to the goal step by step. At each step, you should check:\\
(a) Where the current move leads the player to (the row and column);\\
(b) What is in that grid. Is it a hole? Is it the goal? Is it an empty space?\\
(c) Determine if that is a safe action. If not, correct it and re-generate the action plan.\\
3. Next, verify if the steps successfully navigate the player to the goal without falling into the hole. If not, restart from step 2 and re-generate this step.\\
4. If succeed, output an aggregated plan using "Action plan: <PLAN>", where <PLAN> is a string concatenated action in each step. For example, "Action plan: L,L,R,U,D" meaning an action plan of left, left, right, up, and down. Double check the final action plan is consistent with the previous analysis.\\
Do not output any extra content after the above aggregated output.\\
\\
Please generate the action plan for the following maze:\\
This is a 3x3 map.\\
The player is at: row 3, column 2;\\
There is no holes in this map;\\
The goal is at: Row 1, Column 2.
\end{tcolorbox}

\section{Prompt with in-context examples}\label{prompt-ic-example}
In Section~\ref{visual-vs-text-sec}, we described the procedure of including in-context example in the test. Below, we use the main task in the maze navigation scenario as an example to show the complete prompt after adding in-context examples.
\begin{tcolorbox}[left=1.2pt,right=1.2pt,top=1.2pt,bottom=1.2pt]
\textbf{Prompt for Maze Navigation scenario, Main task (Spatial Planning, in-context examples):}
\\
\\
As a professional maze solver, your task is to analyze a grid-based map and devise an action plan that enables a player to reach the goal from the starting point without falling into any holes, using the fewest possible moves. Since coding is not within your skill set, your approach relies on logical reasoning of the map.
\\
\\
\#\# Game Setup\\
- The game presents a fully observable grid-based map.\\
- The player starts at a specified grid square, with the goal located elsewhere on the map.\\
- Each grid square is either safe or contains a hole.\\
- Your goal is to guide the player to the goal while avoiding holes.\\
The following figure shows how the player, the holes (non-safe grid), the lands (safe grids), and the goals look like.\\
\\
\includegraphics[width=0.3\textwidth]{figures/supp/system-figure-1.png}
\\
\\
\#\# Moving Rules\\
- The action plan involves a series of moves: `L' (left), `R' (right), `U' (up), or `D' (down).\\
- Each move transfers the player to the adjacent square in that direction, provided it is a safe square. The player cannot move more than one square at a time.\\
- Moving off the edge of the map has no effect. The player will remain at the same square.\\
- DO NOT MOVE INTO A HOLE! Falling into a hole results in defeat.\\
- Locating at the grid containing the goal results in victory.\\
We provide an example to further illustrate the rules.\\
\\
\includegraphics[width=0.25\textwidth]{figures/supp/system-figure-2.png}
\\
\\
In this provided example:\\
- The player is at Row 1, Column 1;\\
- The goal is at Row 4, Column 4;\\
- There are two holes: one at Row 1, Column 2, and another at Row 4, Column 1.\\
- The player can move DOWN. This is because moving down brings them to Row 2, Column 1, and this cell is safe (without holes).\\
- Moving UP has no effects. This is because the player is already in the topmost row.\\
- Similarly, moving LEFT has no effects because the player is already in the left-most column.\\
- Moving RIGHT places the player at Row 1, Column 2. Since there is a hole at this grid, this move results in a loss.\\
\textit{(continue in next page)}\\
\end{tcolorbox}

\begin{tcolorbox}[left=1.2pt,right=1.2pt,top=1.2pt,bottom=1.2pt]
\textit{(Continue)}\\
\#\# Procedure and Output\\
Now you will solve the given maze. To solve it, please generate text exactly follow the following steps:\\
1. First, interpret map. List where the player is at now, where is the goal, and where are the holes.\\
2. Then, generate an action plan to navigate to the goal step by step. At each step, you should check:\\
(a) Where the current move leads the player to (the row and column);\\
(b) What is in that grid. Is it a hole? Is it the goal? Is it an empty space?\\
(c) Determine if that is a safe action. If not, correct it and re-generate the action plan.\\
3. Next, verify if the steps successfully navigate the player to the goal without falling into the hole. If not, restart from step 2 and re-generate this step.\\
\\
\#\# Example:\\
\includegraphics[width=0.25\textwidth]{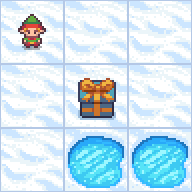}
\\
<Interpret>\\
The player is at row 1, column 1, and the goal is at row 2, column 2.\\
There are 2 holes. They are at: row 3, column 2; row 3, column 3.\\
<Action Plan>\\
- Moving Right (R). The player is now at row 1, column 2. This grid is safe.\\
- Moving Down (D). The player is now at row 2, column 2. This grid is the goal, so we stop here.\\
<Verification>\\
1. Right to row 1, column 2 (safe)\\
2. Down to row 2, column 2 (goal)\\
<Output>\\
Action plan: R,D\\

Please generate the action plan for the following maze:
\\
\\
\includegraphics[width=0.25\textwidth]{figures/supp/maze-t5-eg.png}
\end{tcolorbox}
\newpage
\clearpage

\section{Training Details}\label{training-details}
In this section, we describe the training details when we fine-tune LLaVA and InternLM-XComposer for our designed tasks. We perform LoRA fine-tuning, and we stick with the default hyperparameter settings in their official repo. The detailed hyperparameter choices are shown in Table~\ref{tab:supp-hyperparameters}.

\begin{table*}
    \centering
    \caption{Training details on LLaVA and InternLM-XComposer.}
    \begin{tabular}{lll}
    \toprule
     &  & Value \\
    \midrule
    \multirow{7}{*}{\textbf{LLaVA}} &  Learning rate & 2e-4 \\
        & Scheduler & Cosine \\
        & Epoch & 1 \\
        & Training data & 10k \\
        & Batch size & 32 \\
        & Pretrained Checkpoint &  \texttt{llava-v1.6-vicuna-7b} \\
    \midrule
    \multirow{7}{*}{\textbf{InternLM}} & Learning rate  & 5e-5 \\
        & Scheduler & Cosine \\
        & Epoch & 1 \\
        & Training data & 10k \\
        & Batch size & 8 \\
        & Pretrained Checkpoint &  \texttt{internlm-xcomposer2-7b} \\
    \bottomrule
    \end{tabular}
    \label{tab:supp-hyperparameters}
\end{table*}

\section{Complete Task Performance Results with different difficulty levels}\label{table-different-level}
In this section, we present the experimental results of models across different difficulty levels. The results are shown in Table~\ref{tab:complete-results}. As expected, we observe that as difficulty increases, all models perform progressively worse, with some performing close to random guessing at higher difficulty levels (\emph{e.g.,} Task 1 in Maze Navigation scenario). We also observe that GPT-4o, the most recently released model, performs the best across different tasks, although it still frequently makes mistakes under different difficulty levels. This suggests a current bottleneck in state-of-the-art VLMs.

\begin{table}[t!]
\small
\centering
\caption{Model performance on task 1 - 4 with different difficulties.} 
\label{tab:complete-results}
\resizebox{0.75\textwidth}{!}{
\begin{tabular}{@{}lccccccccc@{}}
\toprule
  \textbf{Task 1} & \multicolumn{6}{c}{\textsc{Maze Navigation}}
       & \multicolumn{3}{c}{\textsc{Blocksword}} \\
     \cmidrule(lr){2-7}
     \cmidrule(lr){8-10}
     Difficulty & \textbf{3} & \textbf{4} & \textbf{5} & \textbf{6} & \textbf{7} & \textbf{8} & \textbf{3} & \textbf{4} & \textbf{5} \\
\midrule
 
 {\small Gemini \cite{team2023gemini}} &   0.63&	0.58&	0.61&	0.45&	0.64 & 	0.54 &  0.86	&0.82	& 0.89 \\
 {\small GPT-Vision \cite{achiam2023gpt}  } & 0.60&	0.56&	0.56&	0.47&	0.62&	0.54& 0.77&	0.70	&0.72  \\ 
 {\small Claude-3 \cite{anthropic2024claude}   } &   0.44&	0.41&	0.39&	0.42&	0.55&	0.49& 0.52&	0.40&	0.36          \\ 
 {\small GPT-4o \cite{GPT-4o}  } & 0.72 &	0.65&	0.57&	0.44&	0.56&	0.53 &0.98&	0.94&	0.92\\
\midrule
  \textbf{Task 2} & \multicolumn{6}{c}{\textsc{Maze Navigation}}
       & \multicolumn{3}{c}{\textsc{Blocksword}} \\
     \cmidrule(lr){2-7}
     \cmidrule(lr){8-10}
     Difficulty & \textbf{3} & \textbf{4} & \textbf{5} & \textbf{6} & \textbf{7} & \textbf{8} & \textbf{3} & \textbf{4} & \textbf{5} \\
\midrule
 
 {\small Gemini \cite{team2023gemini}} &   0.69 &	0.65&	0.53&	0.46&	0.54	&0.47& 0.63	&0.57	&0.32  \\
 {\small GPT-Vision \cite{achiam2023gpt}  } & 0.37	&0.27&	0.22&	0.30&	0.29&	0.18&  0.86&	0.77&	0.77  \\ 
 {\small Claude-3 \cite{anthropic2024claude}   } &   0.65	&0.65	&0.70	&0.65&	0.67&	0.70 &0.59	&0.57&	0.43          \\ 
 {\small GPT-4o \cite{GPT-4o}  } & 0.80 &	0.63&	0.65&	0.64&	0.66&	0.64& 0.90&	0.92&	0.87 \\
\midrule

  \textbf{Task 3} & \multicolumn{6}{c}{\textsc{Maze Navigation}}
       & \multicolumn{3}{c}{\textsc{Blocksword}} \\
     \cmidrule(lr){2-7}
     \cmidrule(lr){8-10}
     Difficulty & \textbf{3} & \textbf{4} & \textbf{5} & \textbf{6} & \textbf{7} & \textbf{8} & \textbf{3} & \textbf{4} & \textbf{5} \\
\midrule
 
 {\small Gemini \cite{team2023gemini}} &   0.38 &	0.32&	0.36&	0.23&	0.37&	0.30& 0.62&	0.51&	0.49  \\
 {\small GPT-Vision \cite{achiam2023gpt}  } &0.79&	0.41&	0.43&	0.37&	0.33&	0.40& 0.68&	0.76&	0.67  \\ 
 {\small Claude-3 \cite{anthropic2024claude}   } &   0.35&	0.22	&0.18&	0.32&	0.35&	0.47 & 0.52	&0.45&	0.49          \\ 
 {\small GPT-4o \cite{GPT-4o}  } &0.89	&0.72&	0.49&	0.4&	0.39&	0.59 &0.85&	0.95&	0.90 \\
\midrule

  \textbf{Task 4} & \multicolumn{5}{c}{\textsc{Maze Navigation}}
       & \multicolumn{4}{c}{\textsc{Blocksword}} \\
     \cmidrule(lr){2-6}
     \cmidrule(lr){7-10}
     Difficulty & \textbf{1} & \textbf{3} & \textbf{5} & \textbf{7} & \textbf{9} & \textbf{1} & \textbf{3} & \textbf{5} & \textbf{7} \\
\midrule
 
 {\small Gemini \cite{team2023gemini}} &   0.47&	0.47&	0.56	&0.49	&0.46& 0.64&	0.57&	0.50	&0.50 \\
 {\small GPT-Vision \cite{achiam2023gpt}  } & 0.62	&0.55&	0.57&	0.52	&0.53 &0.66&	0.70&	0.74&	0.73  \\ 
 {\small Claude-3 \cite{anthropic2024claude}   } &   0.57&	0.60&	0.60&	0.59&	0.67& 0.72	&0.65	&0.60&	0.68          \\ 
 {\small GPT-4o \cite{GPT-4o}  } & 0.72&	0.78	&0.79	&0.67	&0.76 &0.92&	0.73&	0.70&	0.68 \\
\midrule

\end{tabular}
\vspace{-2ex}
}

\end{table}

\section{Benchmark Documentation and Intended Users}\label{assets-quality}
The VSP benchmark is designed to evaluate VLM's capability in visual spatial planning. Visual spatial planning refers to the ability to comprehend the spatial arrangements of objects and devise action plans to achieve desired outcomes in visual scenes. The VSP benchmark consists of two scenarios: \ding{182} the simulated \textbf{Maze Navigation scenario}, whose main task is to move a game character through a maze, and \ding{183} the photo-realistic \textbf{Blocks World scenario}, whose main task is to move blocks from a starting configuration to a goal configuration. In each scenario, in addition to the main task, VSP introduces four sub-tasks that focus on the individual capabilities needed for the main task:

$\bullet$ \textbf{T1. Single Object Perception} -- Determine the characteristics of a single object; 

\vspace{-0.04in}
$\bullet$ \textbf{T2. Spatial Relation Perception} -- Determine the relative positions of two objects; 

\vspace{-0.04in}
$\bullet$ \textbf{T3. Environment Perception} --  Find textual descriptions that describe the visual environment;

\vspace{-0.04in}
$\bullet$ \textbf{T4. Spatial Reasoning} -- Determine the consequence of a series of actions or moves. 

The VSP benchmark consists of 10 tasks in two scenarios. For each task, the problems are designed with different difficulty levels. Specifically, each difficulty level consists of 100 problems. In total, the VSP benchmark includes 4.4k questions. To implement the two scenarios, we utilize and enhance existing resources from two aspects. For the maze navigation scenario, we leverage OpenAI's gym~\cite{brockman2016openai} engine to generate input images. For the blocks world scenario, we sample input images from the BIRD dataset~\cite{gokhale2019blocksworld}. The BIRD dataset is originally built to test an RL model's capability in understanding visual block configurations and performing sequential actions to reach the target state. We enhance it by designing auxiliary tasks (T1-T4), corresponding textual descriptions, and text prompts necessary for the input of VLMs. Additionally, we implement auto-evaluation scripts for each task, aiming to provide a convenient testbed for current VLMs. All the images and texts in this benchmark do not contain any personally identifiable information or offensive content.

All the content in this benchmark can be accessed, reviewed, and downloaded via \url{https://github.com/UCSB-NLP-Chang/Visual-Spatial-Planning}. As the authors of this benchmark, we assume full responsibility for any rights violations related to this benchmark. The benchmark is licensed under the MIT license. We will consistently monitor issues and pull requests for better maintenance. Additionally, we also release the test scripts to replicate our results.

\section{Limitation}\label{limitation}
It is important to note that our proposed VSP benchmark also has limitations.
First, as a VLM benchmark specifically tailored for visual spatial planning capability, the VSP does not measure a VLM's abilities in other important aspects, such as semantic understanding, factual knowledge, \emph{etc.}
We emphasize that VSP is not a comprehensive benchmark for VLMs, but rather a benchmark focusing on an important capability that has been mostly overlooked by existing benchmarks.
Second, we also note that the appearance of objects in the image may influence models' performance. Specifically, a model might find it easier to recognize objects against a darker background in the blocks world scenario, and vice versa. The current measure is based on a single kind of object appearance, which might favor some particular models trained on similar images. An ideal measurement would assess the average performance on images with the same content but a variety of different appearances.
That said, with the detailed prompt description and sufficient information provided in the image, we believe the current version of the VSP benchmark already demonstrates the deficiencies of current state-of-the-art models in visual spatial planning. In future work, we plan to incorporate appearance/style variations in the input images for a more thorough model ability quantification.

\end{document}